\documentclass[runningheads]{llncs}
\usepackage[T1]{fontenc}
\usepackage{graphicx}
\usepackage[numbers,sort&compress]{natbib}

\usepackage{colortbl}

\usepackage[disable]{todonotes}
\usepackage{pifont}
\usepackage{tikz}
\usepackage{listofitems}
\usetikzlibrary{shapes.geometric, arrows, arrows.meta, positioning, fit, math, chains}
\usetikzlibrary[calc]

\tikzstyle{arrow} = [-{Latex[length=3mm]}]

\tikzstyle{above-text} = [text centered, text width=3cm, anchor=south]
\tikzstyle{right-text} = [text centered, text width=3cm, anchor=west]
\tikzstyle{left-text} = [text centered, text width=3cm, anchor=east]

\tikzstyle{main-node} = 
    [rectangle, 
    rounded corners=5pt, 
    minimum width=50pt, 
    minimum height=20pt,
    text centered, 
    draw=black,
    thick]

\tikzstyle{condition-node} = 
    [diamond, 
    minimum width=3cm, 
    minimum height=1cm, 
    text centered, 
    draw=black, 
    text width=1cm]

\usepackage{array}
\usepackage[singlelinecheck=false 
]{caption}

\usepackage[utf8]{inputenc}
\usepackage{bbding}
\usepackage{multirow}
\usepackage{fontawesome5}  

\usepackage[nolist,nohyperlinks]{acronym}
\begin{acronym}
  \acro{ACG}{Automatic Code Generator}
  \acro{LLM}{Large Language Model}
\acro{GTP}{Generative Pre-Trained Transformers}
\acro{CS}{Computer Science}
\acro{TSP}{Traveling Salesman Problem}
\acro{LLM}{Large Language Model}
\acro{AoC}{Advent of Code}
\acro{Copilot}{GitHub Copilot}
\acro{NLP}{Natural Language Processing}
\acro{VSC}{Visual Studio Code}
\acro{CP}{Constraint Programming}
\end{acronym}

\usepackage{paralist}
\usepackage{listing}

\usepackage[many]{tcolorbox}

\usepackage{hyperref}

\usepackage{color}

\definecolor{pblue}{rgb}{0.13,0.13,1}
\definecolor{pgreen}{rgb}{0,0.5,0}
\definecolor{pred}{rgb}{0.9,0,0}
\definecolor{pgrey}{rgb}{0.46,0.45,0.48}
\definecolor{keywordColor}{rgb}{0.000000, 0.000000, 0.635294}
\definecolor{stringColor}{rgb}{0.558215, 0.000000, 0.135316}
\colorlet{numberColor}{red!60!black}

\newcommand{\colbackcolor}{gray!5!white}
\newcommand{\colframecolor}{white!60!black}

\usepackage{listings}

\lstdefinestyle{cstyle}{%
    language=C,
    basicstyle = \ttfamily\small,
    mathescape = true,
    morekeywords = {u32,u8,u16,bool,int,i8,i16},
    keywordstyle=\color{keywordColor}\bfseries,
    numbers=left,
    identifierstyle=\color{black}\ttfamily,
    commentstyle=\itshape\ttfamily\textcolor{commentsColor},
    morekeywords = [2]{min},
    keywordstyle = [2]\color{green!40!black}\ttfamily\bfseries,
}

\lstdefinestyle{boxesstyle}{%
    language=Python,
    basicstyle = \ttfamily\small,
    mathescape = true,
    keywordstyle=\color{keywordColor}\bfseries,
    numbers=left,
    identifierstyle=\color{black}\ttfamily,
    commentstyle=\itshape\ttfamily\textcolor{commentsColor},
    morekeywords = [2]{pack_items},
    keywordstyle = [2]\color{green!60!black}\ttfamily\bfseries,
}

\newlength\defaultparindent
\AtBeginDocument{\setlength\defaultparindent{\parindent}}

\newcommand{\rqtwo}{What is the capability of \ac{LLM} tools at
  identifying the problem to solve?}
\newcommand{\rqone}{How effective are \ac{LLM} tools at solving
  advanced programming assignments correctly?}
\newcommand{\rqthree}{How does the choice of \ac{LLM} tool and programming language affect source-code generation?}

\usepackage{subfig}

\newtcolorbox{promptbox}[2][]
{
  enhanced, 
  breakable,
  skin first=enhanced,
  skin middle=enhanced,
  skin last=enhanced,
  enlarge top by = -2pt,
  top = 2pt,
  bottom = 2pt,
  right = 2pt,
  left = 2pt,
  colframe = black,
  colback  = \colbackcolor,
  title    = {#2},
  #1,
}

\usepackage{etoolbox}

\usepackage{kantlipsum} 

\newcommand{\repthanks}[1]{\textsuperscript{\ref{#1}}}
\makeatletter
\patchcmd{\maketitle}
  {\def\thanks}
  {\let\repthanks\repthanksunskip\def\thanks}
  {}{}
\patchcmd{\@maketitle}
  {\def\thanks}
  {\let\repthanks\@gobble\def\thanks}
  {}{}
\newcommand\repthanksunskip[1]{\unskip{}}
\makeatother

\usepackage{hyperref}

\begin{document}
\title{Evaluating Code Generation of LLMs in Advanced Computer Science Problems}
\titlerunning{Evaluating Advanced Course Assignments with LLMs}
%
\author{Emir Catir\thanks{Both authors contributed equally\protect\label{bothauthors}}\inst{1}\orcidID{0009-0001-6314-5720} \and
Robin Claesson\repthanks{bothauthors}\inst{1}\orcidID{0009-0003-5083-582} \and
Rodothea Myrsini Tsoupidi\inst{2}\orcidID{0000-0002-8345-2752}}
\authorrunning{E. Catir, R. Claesson, and R.M. Tsoupidi}
%
\institute{Royal Institute of Technology, Stockholm, Sweden\\
  \email{\{robcla,catir\}@kth.se}\\
\and
Independent Researcher, Stockholm, Sweden\\
\email{rtsoupidi@acm.org}}
\maketitle              
%

\begin{abstract} 
  Large Language Models (LLMs), such as GitHub Copilot and ChatGPT have become popular among programming students. Students use LLMs to assist them in programming courses, including generating source code.

Previous work has evaluated the ability of LLMs in solving introductory-course programming assignments. The results have shown that LLMs are highly effective in generating code for introductory Computer Science (CS) courses. However, there is a gap in research on evaluating LLMs’ ability to generate code that solves advanced programming assignments.

In this work, we evaluate the ability of four LLM tools to solve programming assignments from advanced CS courses in three popular programming languages, Java, Python, and C. We manually select 12 problems, three problems from introductory courses as the baseline and nine programming assignments from second- and third-year CS courses. To evaluate the LLM-generated code, we generate a test suite of 1000 test cases per problem and analyze the program output.

Our evaluation shows that although LLMs are highly effective in generating source code for introductory programming courses, solving advanced programming assignments is more challenging. Nonetheless, in many cases, LLMs identify the base problem and provide partial solutions that may be useful to CS students. Furthermore, our results may provide useful guidance for teachers of advanced programming courses
on how to design programming assignments.

\end{abstract}

\section{Introduction}
In recent years, advances in machine learning have enabled
high-quality analysis of natural language for diverse purposes,
including chatbots, image, and code generation.
Textual \acp{LLM} are generative machine-learning models that use
large quantities of data during the training process.
Given a prompt, namely a text sequence, the trained model
generates a text response that is likely to follow the provided
prompt.

Many \acp{LLM} have the ability to generate code when the data they
are trained on includes code examples.
\acp{ACG}, such as Github Copilot\footnote{Copilot: \url{https://github.com/features/copilot}}, are specialized
\acp{LLM} that are trained for source-code generation.
Chatbots, such as ChatGPT\footnote{ChatGPT
  \url{https://chat.openai.com/chat}} and Hugging Face
chats\footnote{Hugging Face: \url{https://huggingface.co/chat}} are
general-purpose \acp{LLM} that may be trained on code examples.
Many students use different \ac{LLM} tools, including chatbots and
\acp{ACG} to receive assistance for their programming
assignments~\cite{becker2023programming,kasneci2023chatgpt,keuning2024students,cipriano2024chatgpt,margulieux2024self}.
A survey by \citeauthor{keuning2024students} with 264 responses shows
that the use of \ac{LLM} tools among students has increased in recent
years~\cite{keuning2024students}.
As a response, many teachers have started including \acp{LLM} in their
educational plan or adopt policies to ensure that the students
receive the required education.

Previous research has shown that \ac{LLM} tools are highly capable at
solving introductory programming
tasks~\cite{denny2023conversing,chen2021evaluating,finnie2022robots}
and perform well in introductory \ac{CS}
courses~\cite{avelka2023CanGP}.
In particular, given a prompt that describes the program to implement
in natural language, \ac{LLM} tools can generate functional code after
one or multiple
attempts~\cite{denny2023conversing,chen2021evaluating,finnie2022robots}.

Recent work has shown that advanced programming students believe that
the use of \ac{LLM} tools assists them but they also face problems
such as inaccurate or incorrect answers
\cite{korpimies2024unrestricted,arora2024analyzing,choudhuri2024insights}.
In this work, we investigate how \ac{LLM} tools handle advanced
programming assignments.
In particular, we focus on programming assignments from second and
third-year university courses to investigate how \ac{LLM} tools handle them.
Typically, these problem descriptions aim to train the problem-solving
abilities of the students and describe an everyday-life problem that
the students need to solve.
The question here is whether \ac{LLM} tools can decipher a
natural-language described programming assignment, recognize the
algorithmic problem to solve, and generate functional code that solves
the problem.

More specifically, in this work, we evaluate the ability of a set of
publicly available \ac{LLM}-based tools to generate code for a
selected set of nine programming assignments from advanced \ac{CS}
courses.
We ask the \acp{LLM} to generate code in three popular programming
languages, Java, Python, and C.

We pose the following research questions to guide our research:
\begin{itemize}
\item[\textbf{RQ1:}] \rqone
\item[\textbf{RQ2:}] \rqtwo
\item[\textbf{RQ3:}] \rqthree
\end{itemize}

All evaluation data and  prompts used for generating code are available at
\url{https://github.com/Emir-Catir-and-Robin-Claesson/publish}.

\section{Related Work}

\paragraph{Evaluating \acp{LLM} in \ac{CS} courses.}
In a broad scope, \citeauthor{avelka2023CanGP} evaluate the
performance of GitHub Copilot at solving multiple computer education
tests, including multiple-choice questions and programming exercises
in Python.
Their findings show that while the \ac{ACG} model fails to pass the
course, it achieves high score in most exercises and manages to
improve the answers based on the auto-grader
feedback~\cite{avelka2023CanGP}.
In a different approach, \citeauthor{reeves2023evaluating} evaluate
the performance of Github Copilot at solving Parsons Problems, namely
programs where the program code is given in incorrect
order.
They show that Copilot can solve 80\% of the problems in Python if 
ignoring indentation errors~\cite{reeves2023evaluating}.

\paragraph{Evaluating \acp{LLM} in introductory programming assignments.}
\citeauthor{denny2023conversing} investigate the solving ability of
 GitHub Copilot on a large set of simple programming exercises in
 Python.
 When the \ac{LLM} assistant fails to generate a correct solution, the
 authors try to provide more clear instructions in natural language
 (prompt engineering).
In total, Copilot fails to provide correct solution to 20\% of the
problems~\cite{denny2023conversing}.
\citeauthor{finnie2022robots} assess the ability of GitHub Copilot to
solve introductory programming problems, such as the Rainfall
problem~\cite{soloway1986learning}.
Their work includes evaluating multiple generated programs for a set
of programming tests in a programming course.
Their findings show that GitHub Copilot received high scores on the
 programming tests outperforming most of the
students~\cite{finnie2022robots}.

\paragraph{Evaluating \acp{LLM} in advanced programming assignments.}
\citeauthor{llms_cp_2024} investigate the use of \acp{LLM} for the
automatic transformation of textual problem descriptions into concrete
\ac{CP} specifications\footnote{Constraint programming is a method for
solving combinatorial problems.}.
Their experiments include a set of exercises from a \ac{CP} course and
show that the assisted \ac{LLM} can solve from 65\% for easier
problems up to 35\% for the more complex assignments.
\citeauthor{finnie2023my} investigate the performance of and \ac{ACG},
OpenAI Codex, at solving exams questions of a CS2 course,
\textit{Data structures and Algorithms}.
Their results show that the \ac{ACG} performs better than students
in this course~\citeauthor{finnie2023my}.
Instead, in this paper, we evaluate different \acp{ACG} on more
advanced courses, namely CS4 and CS5.

\section{Motivating Example}
\label{sec:motivating-example}
Consider the \textit{bin-packing} problem, a famous NP-complete problem where the objective is to place a
set of weighted items in the minimum number of bins of specified
capacity.
Figure~\ref{fig:boxes-prompt} shows the high-level description of the bin-packing problem that was given during a lecture in a CS5 course, \textit{Algorithms, Data Structures, and Complexity}. 

\begin{figure}[ht!]
\begin{tcolorbox}[colback=\colbackcolor, colframe=\colframecolor, top = 2pt, bottom = 2pt,
  right = 2pt, left = 2pt]
A person is moving out of their house and need to pack all their belongings into boxes. They have an infinite number of boxes available, but want to use as few as possible.
The person has a list of all their items that need to be packed in boxes. All boxes have the same weight capacity.
\end{tcolorbox}
\caption{\label{fig:boxes-prompt} Problem description prompt for the
  \texttt{boxes} problem (Table~\ref{tab:problems}). }
\end{figure}

We want to investigate, whether \ac{LLM} tools are able to 1) identify
the problem and 2) generate correct code to solve or approximate the
problem.
For example, identifying the problem as the \textit{bin-packing} problem is a
significant assistance to a student.
Apart from identifying the problem, an \ac{LLM} tool may be able to
generate code that solves the problem correctly, in our case, to find the
optimal solution to the bin-packing problem.

When we provided the problem prompt in Figure~\ref{fig:boxes-prompt} to Github Copilot, the tool generated the code in Listing
\ref{listing:bin-packing-python}.
The code defines a function \texttt{pack\_items}, which takes as input
the capacity of the boxes, \texttt{weight\_capacity}, and the list of
\texttt{items} that need to fit in the boxes.
In the generated code snippet, line 1 defines the method, and lines 2-3
declare needed variables.
The generated algorithm (lines 5-15) uses a straightforward
heuristic: all listed items are sequentially assigned into a box
(lines 9 to 10).
When adding an item to the current box surpasses the weight limit
(line 6), the algorithm assigns the item to a new box (line 7 and 8).
This solution is a greedy heuristic and may return the optimal
solution in some cases but does not find the optimal solution in the general case.

It is worth mentioning that we attempted to modify the original
prompt, with clear indications that the tool should use a
\textit{complete} algorithm or use \acf{CP}, a
combinatorial solving technique, with no success.
In the first case, the generated code was again a heuristic, and in the
second case, the generated code was incorrect.
Another interesting result is that all the tested \ac{LLM} tools in
all three languages we tested, Java, Python, and C, produce similar
heuristics.

\begin{listing}
\begin{lstlisting}[style=boxesstyle,
  xleftmargin=\defaultparindent]
def pack_items(weight_capacity, items):
  boxes_needed = 0
  current_box_weight = 0

  for item_weight in items:
    if current_box_weight + item_weight > weight_capacity:
      boxes_needed += 1
      current_box_weight = item_weight
    else:
      current_box_weight += item_weight

  if current_box_weight > 0:
    boxes_needed += 1

  return boxes_needed;
...    
\end{lstlisting}
\caption{Python code generated by Github Copilot for the \texttt{Boxes} problem.} 
\label{listing:bin-packing-python}
\end{listing}

To evaluate the solution in
Listing~\ref{listing:bin-packing-python}, we generated 1000 random
problem instances and compared the answers of the tool-generated
solution with our reference solution.
Out of the 1000 problem instances, the tool-generated code
produces 754 correct answers.
Although the generated solution is not correct, the code is still
interesting because it partially solves the problem and may provide a
significant assistance to students.

To identify similar cases, where an \ac{LLM} has solved part of the
problem, we introduce the measure of \textit{accuracy} that measures
the rate of correctly verified test cases over the total number of
test cases.
We define this metric in Section~\ref{sec:code-verification}.

\section{Methodology}
\label{sec:methodology}

\begin{figure}
    \centering
    \resizebox{\textwidth}{!}{
      \begin{tikzpicture}[align=center]  
      \def\ypos{{0,-20,0,20}}
      \def\xpos{{0,70,70,70}}
      \def\pos{-20,0,20}
      \def\posmore{-30,-10,10,30}
     \def\langs{Java,C,Python}
    \def\llms{{codepal,copilot,hf}}
    \node[label=prompt] (prompt-node) {{\Huge\faTerminal}};
    \node (copilot-node) [main-node, right=40pt of prompt-node] {Copilot};
    \node (codepal-node) [main-node,  above= 40pt of copilot-node] {CodePal};
    \node (hf-node) [main-node, below= 40pt of copilot-node] {HF \includegraphics[height=10pt]{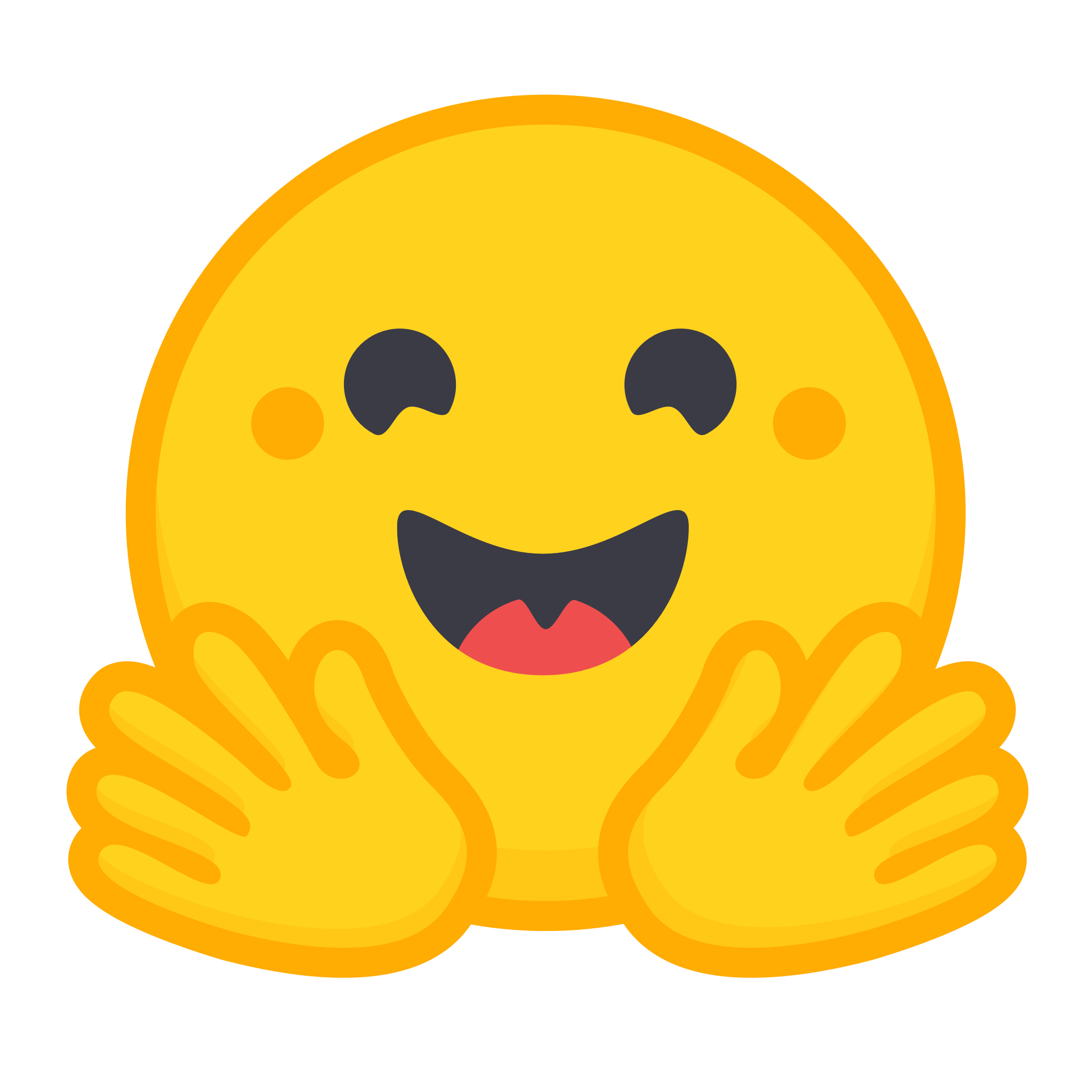}};
    \draw[arrow] (prompt-node) -- (copilot-node);
    \draw[arrow] (prompt-node) -- ($(prompt-node)!0.5!(copilot-node)$) |- (codepal-node);
    \draw[arrow] (prompt-node) -- ($(prompt-node)!0.5!(copilot-node)$) |- (hf-node);

    \foreach [count=\j] \llm in {codepal,copilot,hf}
    {
        \foreach [count=\i] \lang in \langs
        {
           \pgfmathparse{\xpos[\i]}
           \edef\x{\pgfmathresult}
           \pgfmathparse{\ypos[\i]}
           \edef\y{\pgfmathresult}
           \coordinate (mid-point) at ([xshift=\x pt, yshift=\y pt]\llm-node.east);

           \tikzmath{ if \i == 3 || \j == 1  
                   then { \fl = "{\Large\faFileCode[regular]}"; \lng = "\lang"; }
                   else { \fl = ""; \lng = "...";};
           }

           \node[anchor=west] (\llm-node-\i) at (mid-point) {{\fl}{\lng}};
           \draw[arrow] (\llm-node.east) -- ([xshift=30pt]\llm-node.east) |- ([xshift=-10]\llm-node-\i.west);
           }
    }

    \node[main-node,inner sep=5pt,thick,dotted,fit=(prompt-node) (codepal-node) (hf-node) (codepal-node-3) (hf-node-1), label={Code Generation}] {};

    \node[fit=(codepal-node-3) (hf-node-1)] (code-results) {};

    \node (gen-tests-node) [above=30pt of code-results.north] {{\Huge\faCogs}};

    \draw[arrow] (gen-tests-node) -- ($(gen-tests-node)!0.2!(code-results)$) node[right] {1000 tests} -- (code-results);

    \node (verify-node) [main-node,  right=50pt of code-results, minimum height=100pt] {Verifier};

    \foreach \llm in {codepal,copilot,hf}
    {
        \foreach [count=\i] \x in \pos
        {
          \draw[arrow] ([xshift=60pt]\llm-node-\i.west) -| ([xshift=-20pt, yshift=\x]verify-node.west) -- (verify-node);
        }
    }

    \foreach [count=\i] \x in \posmore
    {
        \tikzmath{ if \i == 4
                   then {\ccolor = "green!50!black"; \smb = "100\%"; }
                   else { if \i == 3
                          then {\ccolor = "red!50!black";\smb = "0\%";}
                          else {if \i == 2
                               then {\ccolor = "black";\smb = "...";}
                               else {\ccolor = "orange!30!black";\smb = "20\%";};};};
        }
        \coordinate (m-point) at ([xshift=40pt, yshift=\x]verify-node.east);
        \node[anchor=west] (check-\i) at (m-point) {{\color{\ccolor}\smb}};
        \draw[arrow] ([ yshift=\x]verify-node.east) -- (check-\i);
    }

    \node[main-node,inner sep=2pt,thick,dotted,fit=(code-results) (gen-tests-node) (verify-node) (check-1), label={Code Verification}] (verification-node) {};

\end{tikzpicture}
    }
    \caption{Code Generation and Verification}
    \label{fig:total-flow}
\end{figure}
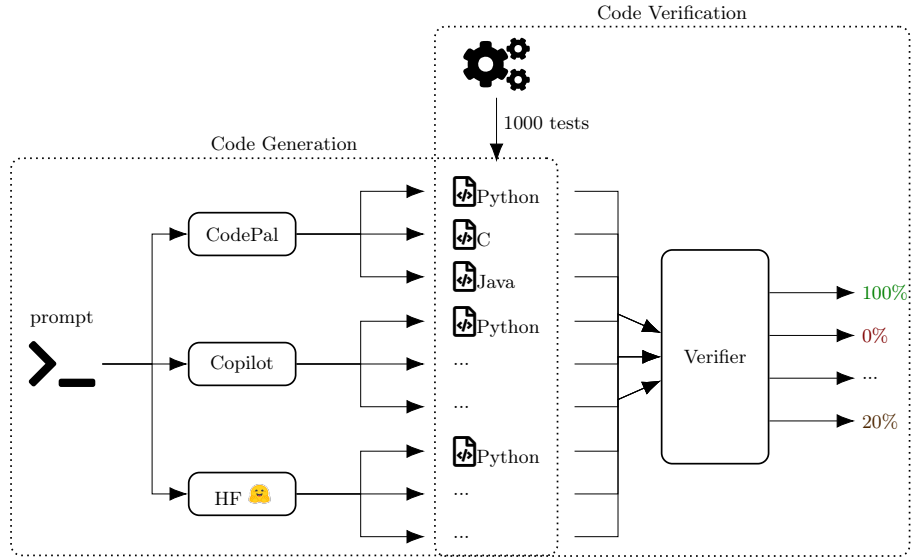

Figure~\ref{fig:total-flow} shows the overview of the methodology,
which consists of two main parts 1) code generation and 2) code
verification part.
The code generation part includes selecting the \ac{LLM} tools
(Section \ref{sec:llm-selection}) and the problems (Section
\ref{sec:problem-selection}) to evaluate.
 For each of the problems, we create a prompt and provide this prompt
 as input to each \ac{LLM} tool to generate source code
 (Section~\ref{sec:code-generation}).

The code verification part includes evaluating the generated solutions
against our reference solutions using a test suite that we generate
per problem (Section~\ref{sec:code-verification}).

\subsection{LLM Selection}
\label{sec:llm-selection}
The selection of \ac{LLM} tools to evaluate is based on the purpose of
this study to evaluate code generation and the tools that are widely accessible to students.
\acp{ACG} are specially trained to generate code, thus the evaluation
uses two \acp{ACG}, Github Copilot and
CodePal\footnote{CodePal: \url{https://codepal.ai}}.
GitHub Copilot is a code-generating tool that integrates with popular IDEs such as Intellij and Visual Studio Code.
CodePal is based on various OpenAI models, mainly on GPT-3.5 and GPT-4, and is a tool that focuses on code
generation.
Students use also chatbots to generate
code~\cite{keuning2024students}.
To investigate general-purpose chatbots, we select two freely
available chatbots in Hugging Face, \textit{Llama3-70b}, with 70
billion parameters, and \textit{Mistral-Nemo-Instruct-2407}, with 12
billion parameters.

\subsection{Problem Selection}
\label{sec:problem-selection}
\begin{table}[!ht]
  \caption{\label{tab:problems} Selected problems for evaluation. \textit{PN}
    stands for the problem number, \textit{Level} shows Computer Science
    course level, and \textit{DA} stands for online description availability. Column \textit{Comments} describes the algorithm of each problem. }
  \centering 
  \begin{tabular}{|c|l|c|c|p{7.5cm}|}
 \hline
PN & Problem         &Level  & DA        & Comments \\\hline
P1 & Temperature     &CS1    & \ding{51} & Rainfall problem~\cite{soloway1986learning} variant\\
P2 & Sorting         &CS1    & \ding{55} & Sorting of string and numeric values\\
P3 & Shortest Path   &CS1    & \ding{51} & Minimum path via intermediate stations\\
P4 & Boxes           &CS5    & \ding{55} & Bin-packing problem\\
P5 & TSP             &CS5    & \ding{55} & Traveling salesman problem\\
P6 & Outlets         &CS5    & \ding{55} & Minimum spanning tree\\
P7 & Cow Game        &CS5    & \ding{51} & Shortest path in grid with obstacles\\
P8 & Dice            &CS5    & \ding{51} & Shortest path in graph given allowed moves \\
P9 & Traveling       &CS5    & \ding{51} & All shortest paths with constraints\\
P10& TV-Zapping      &CS5    & \ding{51} & Minimum increment to match inputs with constraints\\
P11& Cut Boards      &CS4    & \ding{51} & Dynamic Programming\\
P12& Train Shunting  &CS4    & \ding{51} & Similar to Tower of Hanoi\\
\hline 
\end{tabular}
\end{table}

We handpicked programming assignments and problem descriptions from
two \ac{CS} degree programs.
Table~\ref{tab:problems} shows an enumeration of the problems (first
column), a short name describing the problem (second column), the
\ac{CS} level course that the problem corresponds to (third column),
and finally, whether the description is available online (forth
column).
The fifth column provides a brief description of the underlying algorithm.

First, we selected three problems at the introductory \ac{CS} level as
the baseline.
Two of the problems, P1 and P3 are lab assignments from a CS1 course,
\textit{Programming I} at KTH Royal Institute of Technology in Sweden.
P1, \texttt{Temperature}, is a variant of the Rainfall
problem~\cite{soloway1986learning}.
P3
To minimize the chance that the \ac{LLM} that we use is trained on the
actual problem solutions, we introduced an additional problem that was
handwritten by two of the authors of this paper and consists of
sorting a list of people.

Subsequently, we selected a set of nine problems in advanced
programming courses, in particular, CS4 and CS5 courses.
The selected problems have clear textual descriptions that allow the
selected \acp{LLM} to find solutions.
Three of the problems, \texttt{Boxes}, \texttt{TSP}, and
\texttt{Electrical Outlets} are part of the teaching material,
including lecture notes and assignments, in a CS5 course,
\textit{Algorithms, Data Structures, and Complexity} taught at
       KTH Royal Institute of Technology.
       \texttt{Boxes} is an instance of the \textit{bin-packing}
       problem (see Section~\ref{sec:motivating-example}), \texttt{TSP} is an
       instance of the traveling-salesman problem, and
       \texttt{Outlets} is a minimum-spanning tree
       problem.

Four of the problems, the \texttt{Cow Game}, \texttt{Dice},
\texttt{Traveling}, and \texttt{TV-Zapping} are taught in a CS5 course
on \textit{Programming Languages I} at the National Technical
University of Athens in Greece.
The \textit{TSP} was rewritten as a teacher handing out tests in a
classroom rather than a salesman traveling between cities, as it is a
classic CS problem with many available implementations.
\texttt{Cow Game} is a path-finding problem in a grid under
time-induced constraints.
\texttt{Dice} is a shortest-path graph problem with the constraint of
each move having to use specific predefined dice values.
\texttt{Traveling} is a shortest-path problem from a source to all
destinations given that the last edge of the shortest path to each
destination becomes unusable.
\texttt{TV-Zapping} is the problem of converging a number of values
(channels) to the same value (channel) given the constraint than one
channel cannot change two consecutive times.
For these problems the assignment descriptions are available online,
but we were not able to find any of the solutions in GitHub or other
repositories.
It is worth noting that the teachers of this course specify that the
students should not upload their solutions online.

The last two problems, \texttt{Cut Boards} and \texttt{Train
  Shunting}, are problems that are part of a CS4 course,
\textit{Programming II} at KTH.
\texttt{Cut Boards} is the problem of finding the optimal cutting
order of a board in smaller parts with goal to minimize the cutting
cost. The problem is designed to be solved with dynamic programming,
but can also be solved with a greedy algorithm.
\texttt{Train Shunting} is a problem of performing a set of moves to
change the order of wagons in a train. Given three stations, a number
of wagons and their initial and desired order, as well as a
description of the allowed moves, the task is to rearrange the train
by only performing valid moves. The problem is similar to the famous
Tower of Hanoi problem, and the constraint lies in rearranging the
train wagon order, only using valid moves.

\subsection{Code Generation}
\label{sec:code-generation}
The next step in the evaluation is creating the prompt text from the
selected problems and requesting the \ac{LLM} tools to generate code.

The prompts for every problem description (\ac{LLM} prompt) consist of
three parts.
The first part is the problem description, explaining the problem that
the generated code should solve.
Figure~\ref{fig:boxes-prompt} shows the problem description for the
\texttt{Boxes} problem.
The second part describes how the input text for a problem instance is
formatted.
For the \texttt{Boxes} problem the input part is the following:

\begin{tcolorbox}[colback=\colbackcolor, colframe=\colframecolor, top = 2pt, bottom = 2pt,
  right = 2pt, left = 2pt]
The input will be given to standard input in this order: The first row contains the weight capacity of the boxes. The second row contains the number of items. The following rows contain the weight of each item.
\end{tcolorbox}

The last part describes the expected output for a problem
instance. Standardized input and output for the generated problems
allow for automating the testing of the code.

\begin{tcolorbox}[colback=\colbackcolor, colframe=\colframecolor, top = 2pt, bottom = 2pt,
  right = 2pt, left = 2pt]
The output should be printed to standard output in this order: The number of boxes needed to carry all the items.
\end{tcolorbox}

Figure~\ref{fig:total-flow} shows the code generation part in the left
most box.
We provide the three-part prompt as input to the selected \acp{LLM},
GitHub Copilot (Copilot in Figure~\ref{fig:total-flow}), CodePal, and
two models, Llama3-70b and Mistral-Nemo-Instruct-2407, from Hugging
Face (HF in Figure~\ref{fig:total-flow}).
We ask each of the \acp{LLM} to generate code in Python, Java, and C.

Note, that we applied some minor changes to the \acp{LLM}-generated
solutions to run through the verification
process. Table~\ref{tab:table_changes} in
Appendix~\ref{appendix:changed} lists all the generated solutions that
required changes and the change they required. None of these changes
alters the algorithm of the generated code.

\subsection{Code Verification}
\label{sec:code-verification}
Figure~\ref{fig:total-flow} shows the code verification part of the
evaluation (right-most box).
Given the generated programs by the four \ac{LLM} tools, we generate
1000 random test cases for each of the twelve problems.
The test cases for each of the \ac{LLM} solutions use the same random
seed.
We run each \ac{ACG}-generated program separately for each test case,
and the answer given by the program is recorded and verified using our
reference solution, the Verifier (Figure~\ref{fig:total-flow}).
From the collected answers, we report how many test cases were correct
for each generated program and how many cases where wrong.
The output of the Verifier is the \textit{accuracy} of the generated
code based on Definition~\ref{def:accuracy}.
%
\begin{definition}[Accuracy]
\label{def:accuracy}
 Accuracy is the number of correctly solved test cases $s$, divided by
 the total number of test cases, $t$ expressed as $Accuracy =
 \frac{s}{t}$. In this experiment $t = 1000$.
\end{definition}

In the \texttt{Boxes} problem (Section~\ref{sec:motivating-example}),
the accuracy of the GitHub Copilot solution is 75.4\%.
The structure of the verification algorithm varies depending on the
type of problem.
Certain problems have a single correct answer.
For these problems, the verification consists of calculating the
correct answer from the input data and comparing the \ac{LLM} result
to the correct answer.
To account for rounding errors in floating point results, the Verifier
accepts any answer up to one decimal point far from the correct answer
as correct.
Other problems have many possible valid answers, such as the \ac{TSP}
problem.
For these problems, our verification step checks each aspect of the
given answer, such as verifying all nodes exist in a path and that
bounds are not exceeded.
The verification algorithm that checks the \texttt{Boxes} problem (see
Section~\ref{sec:motivating-example}), uses an exhaustive search to check the
solutions that \acp{LLM} generate.

We define a problem correctly solved, when the accuracy is 100\%.

\begin{definition}[Correctness]
\label{def:correctness}
A problem is \textbf{correct} when the accuracy is 100\%, namely $Accuracy = 1$.
\end{definition}

\section{Evaluation}
\label{chap:results}


%



\newcommand{\correct}{\cellcolor{green!25}100}
\newcommand{\error}[1]{\cellcolor{gray!25}#1}
\newcommand{\almost}[1]{\cellcolor{yellow!25}#1}
\newcommand{\wrong}[1]{\cellcolor{red!25}#1}
\newcommand{\some}[1]{\cellcolor{orange!25}#1}

\begin{table*}[!ht]
  \caption{\label{tab:table_all} Evaluation results for each of the
    problems, \ac{LLM} tools, and programming languages. GC stands for
    Github Copilot, CP stands for CodePal, HF stands for Hugging Face,
    LL stands for Llama, and MS stands for Mistral.  RE stands for
    run-time error, IL stands for infinite loop, and CE stands for compilation error.  All values are in \%.}  \centering
  \setlength\tabcolsep{5pt}
  \begin{tabular}{|l|c|c|c|c|c|c|c|c|c|c|c|c|}
 \hline
 \multirow{3}{*}{PN} &  \multicolumn{4}{c}{Python} & \multicolumn{4}{|c}{Java} & \multicolumn{4}{|c|}{C} \\
 \cline{2-13}
 & \multirow{2}{*}{GC} & \multirow{2}{*}{CP} &  \multicolumn{2}{c|}{HF} & \multirow{2}{*}{GC} & \multirow{2}{*}{CP} &  \multicolumn{2}{c|}{HF} & \multirow{2}{*}{GC} & \multirow{2}{*}{CP} &  \multicolumn{2}{c|}{HF} \\
  \cline{4-5}\cline{8-9}\cline{12-13}
  & &  & LL & MS&  &  & LL  & MS  &  & & LL & MS\\\hline


P1& 
\correct & \correct & \correct &\error{RE} & 
\almost{96.9} & \error{RE} & \correct & \almost{96.9} &
\correct & \correct & \correct & \almost{96.9} \\

P2& 
\correct & \correct & \correct & \correct &
\correct & \wrong{0} & \correct & \wrong{0} &
\correct & \correct & \correct & \wrong{0} \\

P3& 
\correct & \correct & \correct & \correct &
\correct & \correct & \correct & \correct &
\correct & \correct & \correct & \correct \\

P4& 
\almost{75.4} & \almost{71.2} & \almost{71.2} & \some{32.3} &
\almost{71.2} & \almost{71.2} & \almost{71.2} & \some{32.3} &
\almost{75.4} & \almost{75.4} & \almost{71.2} & \error{CE}\\

P5& 
\correct & \wrong{0} & \error{RE} & \wrong{0} &
\correct & \wrong{0}& \error{CE} & \wrong{0} &
\wrong{0} & \wrong{0} & \wrong{0} & \error{CE}\\

P6&
\wrong{0} & \wrong{0} & \wrong{0} & \error{RE} & 
\wrong{0} & \error{RE} & \wrong{0} & \error{RE} & 
\wrong{0} & \wrong{0} & \wrong{0} &\error{CE}\\

P7& 
\error{IL} & \wrong{0} & \wrong{0} & \wrong{7.4} &
\wrong{1.7} & \some{51.9} & \wrong{0} & \wrong{7.4} &
\almost{77.4} & \wrong{0} & \wrong{0} & \wrong{2.1}\\

P8& 
\some{23.7} & \some{23.7} & \wrong{6.3} & \error{RE} &
\some{23.7} & \some{23.7} & \wrong{6.3} & \error{RE} &
\some{23.7} & \error{RE} & \error{RE} &\error{CE} \\

P9&
\wrong{0} & \some{13.9} & \wrong{0} & \wrong{0} &
\wrong{1.1} & \wrong{0} & \wrong{0} & \error{RE} &
\some{13.9} & \error{CE} & \error{RE} & \error{CE}\\

P10&
\wrong{1.3} & \almost{96.4} & \wrong{1.4} & \error{RE} &
\wrong{1.4} & \wrong{1.4} & \wrong{1.4} & \wrong{0.3} &
\almost{96.4} & \wrong{3.0} & \wrong{1.4} & \error{CE}\\

P11& 
\correct & \wrong{0} & \wrong{0} & \wrong{0} &
\wrong{0} & \wrong{0} & \wrong{0} & \wrong{0} &
\wrong{0.1} & \wrong{0} & \wrong{0} & \error{CE}\\

P12&
\wrong{0} & \error{IL} & \error{IL} & \wrong{0} &
\wrong{0} & \error{IL} & \error{CE} & \wrong{0} &
\wrong{0} & \wrong{0} & \error{CE} & \wrong{0}\\ 

\hline
\end{tabular}
\end{table*}

This section presents the results of the evaluation of the selected
\ac{LLM} tools (see Table~\ref{tab:problems}).
Table~\ref{tab:table_all} summarizes the results of the evaluation.
For each of problems, P1 to P12, Table~\ref{tab:table_all} shows the
accuracy (see Definition~\ref{def:accuracy}) of the generated source
code for each of the targeted programming languages and each of the \ac{LLM}
tools.
To evaluate \textbf{RQ1}, we compare the results that are
correct in Table~\ref{tab:table_all}.
To evaluate \textbf{RQ2}, we examine how good are the \ac{LLM}
tools at recognizing the algorithm or parts of the algorithm for each
problem.
To evaluate \textbf{RQ3}, we compare the results for the different
programming languages anad \ac{LLM} tools.

\subsection{RQ1: \rqone}
\label{sec:rq1}
The results in Table~\ref{tab:table_all} show the accuracy of each
generated problem.
The problems that are \textit{correct} (see
Definition~\ref{def:correctness}) are colored green in
Table~\ref{tab:table_all}.
The results in Table~\ref{tab:table_all} show that there is a clear
difference between the baseline CS1 problems, P1 to P3, and the
problems from more advanced courses, P4 to P12.
In particular, for P1, we have six out of the twelve solutions
correct, and three of them have accuracy 96.9\%.
The lower accuracy in the last three cases depends on wrong usage of
the minimum value for \texttt{double} numbers in Java and C, where the
value is 0 and not \texttt{-inf} as the algorithm requires.
For problem P2, which sorts a combination of strings and integers, the
failed cases depend on input read mistakes for CodePal and Mistral in Java and
wrong sorting order for Mistral in C.
Problem P3 is correct for all \acp{LLM}.

Among P4 to P12, only P4, \texttt{TSP} and P11, \texttt{Cut Boards},
have two and one correct results, respectively.
In all these cases, Github Copilot is the \ac{LLM} that generates
correct results.
For the \texttt{Cow Game} problem, Github Copilot's solution for C
gives accuracy 77.4\%.
Listing~\ref{lst:cow-error} shows the error of the generated solution
and Listing~\ref{lst:cow-correct} shows the corrected solution.
This error depends on wrong intitialization of the input data.
In particular, assignment \lstinline[style=cstyle,basicstyle=\ttfamily\scriptsize]{grid[x-1][y] = t;} that sets
the time unit at which the specific grid position is marked should
instead be \lstinline[style=cstyle,basicstyle=\ttfamily\scriptsize]{grid[x-1][y] = min(grid[x-1][y], t);}
to consider possible previous markings.

\begin{figure}[!h]
  \hspace{8pt}
  \subfloat[][\label{lst:cow-error} Generated Code]{
\begin{lstlisting}[style=cstyle,basicstyle=\ttfamily\scriptsize]
for (i = 0; i < n; i++) {
  ...
  grid[x][y] = t 
  if (x > 0)
    grid[x-1][y] = t;
  ...
}
\end{lstlisting}}
  \hfill
  \subfloat[][\label{lst:cow-correct} Corrected Code]{
\begin{lstlisting}[style=cstyle, basicstyle=\ttfamily\scriptsize]
for (i = 0; i < n; i++) {
  ...
  grid[x][y] = min(grid[x][y], t);
  if (x > 0)
    grid[x-1][y] = min(grid[x-1][y], t);
  ...
}
\end{lstlisting}}
  \caption{\label{lst:cow}\texttt{Cow Game}: Generated code by Github Copilot in C and corrected code. }
\end{figure}

\paragraph{Summary}
The ability of \acp{LLM} to generate correct solutions in advanced CS
problems compared to CS1 problems is low.
Among all \acp{LLM} Github Copilot is the only tool that is able to
generate correct results.

\subsection{RQ2: \rqtwo}
This research question investigates the ability of \ac{LLM} tools to
identify the problem that the prompt describes or provide useful
information and partial solutions to students.
In Table~\ref{tab:table_all}, we focus on the results that have
partially correct answers or accuracy lower than 100\% but higher than
0\%.

For problem P4, the \texttt{Boxes} problem (see
Section~\ref{sec:motivating-example}), all \ac{LLM} solutions in
Table~\ref{tab:table_all} are implementations of heuristics that approximate the result.
The \acp{LLM} seem to recognize the problem but only provides
suboptimal greedy solutions rather than following the prompt
instructions.

When examining the solutions for P6, the \texttt{Outlets} problem, we
can see that some solutions have functions and comments that 
refer to minimum spanning tree (MST), which is the correct algorithm to
solve this problem. Copilot's solutions in Python and C both reference MSTs and all three solutions generated by Llama  contain a function named \textit{prim}, a reference to Prim's algortihm, which finds the MST.
This indicates that these solutions have correctly identified the
underlying problem.
However, none of them is able to solve any instance of the problem
correctly.
%
We found that this was mostly due to the specific constraints of the tree's intended root as specified in the prompt. 

Problem P7, the \texttt{Cow Game} problem, has two solutions, one from
Github Copilot and the other from CodePal with accuracy
above 50\%.
The Java solution from CodePal finds the shortest path to the
destination cell in a grid, without considering the obstacles.
In Github Copilot's C solution, the algorithm is correct, but when
reading the input code, the implementation overwrites obstacles (see
Listing~\ref{lst:cow-error}).

Is P8, the \texttt{Dice} problem, seven solutions achieve an non-zero
accuracy $>0$ and five solutions achieve 23.7\% accuracy.
The latter solutions find the shortest path to the destination node in
a graph without considering the constraints implied by the
\texttt{dice}, namely the allowed steps.

Problem P9, \texttt{Traveling} problem has two solution that achieve
13.9\% accuracy.
Here, both solutions implement the same algorithm that finds the
second shortest paths.
However, the problem requests for the shortest path after removing one
edge of the initial graph that belongs to the original shortest path.
Similar to the previous cases, here, the two \acp{LLM} recognize the
algorithm but ignore some more problem specific constraints and
instructions.

Problem P10, \texttt{TV-Zapping} problem, includes two solutions
that have a very high accuracy, 96.4\%.
These solutions solve the main problem, but miss a constraint given in
the prompt.
That is, it is not allowed to increase one of the inputs two times subsequently.

Problem P11, \texttt{Cut Board}, was an exercise in dynamic programming.
Almost all \acp{LLM} try to solve the problem using dynamic
programming, but fail to generate correct code to achieve this. The
only solution that is correct uses a greedy algorithm.

Finally, in problem P12, or \texttt{Train Shunting}, which is similar
to \texttt{The Towers of Hanoi} problem, none of the \acp{LLM}
solves the problem and many \acp{LLM} generate solutions that
result in infinite loops or compilation errors.

\paragraph{Summary} The evaluation shows that many \acp{LLM} generate solutions that use heuristics instead of full solutions, solve the problem ignoring some constraints, or recognize the algorithm but fail in the implementation (e.g. Dynamic Programming).
This is an insight that may be useful for teachers that want to
design problems that require understanding and personal effort from
the students.

\subsection{RQ3: \rqthree}
The choice of language affects the result to some extent. In Table \ref{tab:table_all}, \ac{LLM} tools fail to generate functional C code.
In particular, 12 solutions result in either compilation or runtime
errors. However, Github Copilot (GC) provides partial solutions in C for P7
to P10.
Python and Java have eight solutions each that result in errors, and
different \acp{LLM} produce non-runnable code for P1.
In terms of generating partly correct solutions for advanced problems, P4 - P12, we look at solutions with accuracy $>5\%$. \acp{LLM} produced twelve solutions with Python, ten with Java, and seven with C that partially solved the problems.

To compare the \ac{LLM} tools, we see in Table \ref{tab:table_all} that GC performs the best, followed by CodePal (CP). 
Out of 36 solutions, GC produces only a single non-functional
solution, this is for P7, where GC's Python solution results in an
infinite loop.
Solutions produced by CP are failing to a larger extent, while Llama and Mistral generate a larger number or non-functional solutions.
Another interesting insight is that it seems that CP in Python
generates the same heuristic as CG in C, for problems P8 - P10.

\section{Limitations and Future Work}
%
%
In this paper, we have not performed extensive prompt engineering but rather
used the original or slightly modified exercises as provided in the
respective course.
It is possible that prompt engineering may improve the results of
\acp{LLM} in advanced courses.
We leave this as a future work.

The development and improvement of \ac{LLM} tools is constant, which
requires continuous effort to evaluate their capabilities and the
effect of their use on \ac{CS} education.
We believe that there is a need to adjust \ac{CS} course assessments
for a fair assessment of the students' knowledge.

\section{Conclusion}

In this paper, we investigate the capabilities of \ac{LLM} tools to
solve advanced programming assignments.
Our results indicate that compared to introductory course assignments,
\ac{LLM} tools struggle to generate correct source code for advanced
assignments.
However, the tools are often able to recognize the algorithm needed to
solve the assignment.
Our analysis on partially correct results shows that \acp{LLM} have
difficulty adjusting known algorithms to specific constraints and
often tend to implement popular heuristics instead of following the
prompt instructions.
We believe that teachers may use these insights to assess the knowledge
of students in advanced \ac{CS} courses by incorporating variations or constraints to known algorithms in advanced programming assignments.

\bibliographystyle{splncs04nat}
\bibliography{references}

\begin{thebibliography}{16}
\providecommand{\natexlab}[1]{#1}
\providecommand{\url}[1]{\texttt{#1}}
\providecommand{\urlprefix}{URL }
\expandafter\ifx\csname urlstyle\endcsname\relax
  \providecommand{\doi}[1]{doi:\discretionary{}{}{}#1}\else
  \providecommand{\doi}{doi:\discretionary{}{}{}\begingroup
  \urlstyle{rm}\Url}\fi

\bibitem[{Arora et~al.(2024)Arora, Venaik, Singh, Goyal, Tyagi, Goel, Singhal,
  and Kumar}]{arora2024analyzing}
Arora, C., Venaik, U., Singh, P., Goyal, S., Tyagi, J., Goel, S., Singhal, U.,
  Kumar, D.: Analyzing llm usage in an advanced computing class in india. arXiv
  preprint arXiv:2404.04603  (2024)

\bibitem[{Becker et~al.(2023)Becker, Denny, Finnie-Ansley, Luxton-Reilly,
  Prather, and Santos}]{becker2023programming}
Becker, B.A., Denny, P., Finnie-Ansley, J., Luxton-Reilly, A., Prather, J.,
  Santos, E.A.: Programming is hard-or at least it used to be: Educational
  opportunities and challenges of ai code generation. In: Proceedings of the
  54th ACM Technical Symposium on Computer Science Education V. 1, pp. 500--506
  (2023)

\bibitem[{Chen et~al.(2021)Chen, Tworek, Jun, Yuan, de~Oliveira~Pinto, Kaplan,
  Edwards, Burda, Joseph, Brockman, Ray, Puri, Krueger, Petrov, Khlaaf, Sastry,
  Mishkin, Chan, Gray, Ryder, Pavlov, Power, Kaiser, Bavarian, Winter, Tillet,
  Such, Cummings, Plappert, Chantzis, Barnes, Herbert-Voss, Guss, Nichol,
  Paino, Tezak, Tang, Babuschkin, Balaji, Jain, Saunders, Hesse, Carr, Leike,
  Achiam, Misra, Morikawa, Radford, Knight, Brundage, Murati, Mayer, Welinder,
  McGrew, Amodei, McCandlish, Sutskever, and Zaremba}]{chen2021evaluating}
Chen, M., Tworek, J., Jun, H., Yuan, Q., de~Oliveira~Pinto, H.P., Kaplan, J.,
  Edwards, H., Burda, Y., Joseph, N., Brockman, G., Ray, A., Puri, R., Krueger,
  G., Petrov, M., Khlaaf, H., Sastry, G., Mishkin, P., Chan, B., Gray, S.,
  Ryder, N., Pavlov, M., Power, A., Kaiser, L., Bavarian, M., Winter, C.,
  Tillet, P., Such, F.P., Cummings, D., Plappert, M., Chantzis, F., Barnes, E.,
  Herbert-Voss, A., Guss, W.H., Nichol, A., Paino, A., Tezak, N., Tang, J.,
  Babuschkin, I., Balaji, S., Jain, S., Saunders, W., Hesse, C., Carr, A.N.,
  Leike, J., Achiam, J., Misra, V., Morikawa, E., Radford, A., Knight, M.,
  Brundage, M., Murati, M., Mayer, K., Welinder, P., McGrew, B., Amodei, D.,
  McCandlish, S., Sutskever, I., Zaremba, W.: Evaluating large language models
  trained on code. arXiv preprint arXiv:2107.03374  (2021)

\bibitem[{Choudhuri et~al.(2024)Choudhuri, Ramakrishnan, Chatterjee,
  Trinkenreich, Steinmacher, Gerosa, and Sarma}]{choudhuri2024insights}
Choudhuri, R., Ramakrishnan, A., Chatterjee, A., Trinkenreich, B., Steinmacher,
  I., Gerosa, M., Sarma, A.: Insights from the frontline: Genai utilization
  among software engineering students. arXiv preprint arXiv:2412.15624  (2024)

\bibitem[{Cipriano and Alves(2024)}]{cipriano2024chatgpt}
Cipriano, B.P., Alves, P.: "chatgpt is here to help, not to replace
  anybody"--an evaluation of students' opinions on integrating chatgpt in cs
  courses. arXiv preprint arXiv:2404.17443  (2024)

\bibitem[{Denny et~al.(2023)Denny, Kumar, and Giacaman}]{denny2023conversing}
Denny, P., Kumar, V., Giacaman, N.: Conversing with copilot: Exploring prompt
  engineering for solving cs1 problems using natural language. In: Proceedings
  of the 54th ACM Technical Symposium on Computer Science Education V. 1, pp.
  1136--1142 (2023)

\bibitem[{Finnie-Ansley et~al.(2022)Finnie-Ansley, Denny, Becker,
  Luxton-Reilly, and Prather}]{finnie2022robots}
Finnie-Ansley, J., Denny, P., Becker, B.A., Luxton-Reilly, A., Prather, J.: The
  robots are coming: Exploring the implications of openai codex on introductory
  programming. In: Proceedings of the 24th Australasian Computing Education
  Conference, pp. 10--19 (2022)

\bibitem[{Finnie-Ansley et~al.(2023)Finnie-Ansley, Denny, Luxton-Reilly,
  Santos, Prather, and Becker}]{finnie2023my}
Finnie-Ansley, J., Denny, P., Luxton-Reilly, A., Santos, E.A., Prather, J.,
  Becker, B.A.: My ai wants to know if this will be on the exam: Testing
  openai’s codex on cs2 programming exercises. In: Proceedings of the 25th
  Australasian Computing Education Conference, pp. 97--104 (2023)

\bibitem[{Kasneci et~al.(2023)Kasneci, Se{\ss}ler, K{\"u}chemann, Bannert,
  Dementieva, Fischer, Gasser, Groh, G{\"u}nnemann, H{\"u}llermeier
  et~al.}]{kasneci2023chatgpt}
Kasneci, E., Se{\ss}ler, K., K{\"u}chemann, S., Bannert, M., Dementieva, D.,
  Fischer, F., Gasser, U., Groh, G., G{\"u}nnemann, S., H{\"u}llermeier, E.,
  et~al.: Chatgpt for good? on opportunities and challenges of large language
  models for education. Learning and individual differences \textbf{103},
  102274 (2023)

\bibitem[{Keuning et~al.(2024)Keuning, Alpizar-Chacon, Lykourentzou, Beehler,
  K{\"o}ppe, de~Jong, and Sosnovsky}]{keuning2024students}
Keuning, H., Alpizar-Chacon, I., Lykourentzou, I., Beehler, L., K{\"o}ppe, C.,
  de~Jong, I., Sosnovsky, S.: Students' perceptions and use of generative ai
  tools for programming across different computing courses. In: Proceedings of
  the 24th Koli Calling International Conference on Computing Education
  Research, pp. 1--12 (2024)

\bibitem[{Korpimies et~al.(2024)Korpimies, Laaksonen, and
  Luukkainen}]{korpimies2024unrestricted}
Korpimies, K., Laaksonen, A., Luukkainen, M.: Unrestricted use of llms in a
  software project course: Student perceptions on learning and impact on course
  performance. In: Proceedings of the 24th Koli Calling International
  Conference on Computing Education Research, pp. 1--7 (2024)

\bibitem[{Margulieux et~al.(2024)Margulieux, Prather, Reeves, Becker,
  Cetin~Uzun, Loksa, Leinonen, and Denny}]{margulieux2024self}
Margulieux, L.E., Prather, J., Reeves, B.N., Becker, B.A., Cetin~Uzun, G.,
  Loksa, D., Leinonen, J., Denny, P.: Self-regulation, self-efficacy, and fear
  of failure interactions with how novices use llms to solve programming
  problems. In: Proceedings of the 2024 on Innovation and Technology in
  Computer Science Education V. 1, pp. 276--282 (2024)

\bibitem[{Michailidis et~al.(2024)Michailidis, Tsouros, and
  Guns}]{llms_cp_2024}
Michailidis, K., Tsouros, D., Guns, T.: {Constraint Modelling with LLMs Using
  In-Context Learning}. In: Shaw, P. (ed.) 30th International Conference on
  Principles and Practice of Constraint Programming (CP 2024), Leibniz
  International Proceedings in Informatics (LIPIcs), vol. 307, pp. 20:1--20:27,
  Schloss Dagstuhl -- Leibniz-Zentrum f{\"u}r Informatik, Dagstuhl, Germany
  (2024), ISBN 978-3-95977-336-2, ISSN 1868-8969,
  \doi{10.4230/LIPIcs.CP.2024.20},
  \urlprefix\url{https://drops.dagstuhl.de/entities/document/10.4230/LIPIcs.CP.2024.20}

\bibitem[{Reeves et~al.(2023)Reeves, Sarsa, Prather, Denny, Becker, Hellas,
  Kimmel, Powell, and Leinonen}]{reeves2023evaluating}
Reeves, B., Sarsa, S., Prather, J., Denny, P., Becker, B.A., Hellas, A.,
  Kimmel, B., Powell, G., Leinonen, J.: Evaluating the performance of code
  generation models for solving parsons problems with small prompt variations.
  In: Proceedings of the 2023 Conference on Innovation and Technology in
  Computer Science Education V. 1, pp. 299--305 (2023)

\bibitem[{Soloway(1986)}]{soloway1986learning}
Soloway, E.: Learning to program= learning to construct mechanisms and
  explanations. Communications of the ACM \textbf{29}(9), 850--858 (1986)

\bibitem[{\v{S}avelka et~al.(2023)\v{S}avelka, Agarwal, Bogart, Song, and
  Sakr}]{avelka2023CanGP}
\v{S}avelka, J., Agarwal, A., Bogart, C., Song, Y., Sakr, M.F.: Can generative
  pre-trained transformers (gpt) pass assessments in higher education
  programming courses? Proceedings of the 2023 Conference on Innovation and
  Technology in Computer Science Education V. 1  (2023)

\end{thebibliography}
\appendix
\section{Changed solutions}
\label{appendix:changed}

We made some small alterations to the \ac{LLM}-generated solutions before the tests. Changes to a solution depend on one of the following three reasons, 1) to make the program running, 2) the generated solution contained a running algorithm but required minor changes to the output to pass the verification algorithm, and 3) to lower the runtime of the programs. No changes were made to the algorithms of the solutions.

Table \ref{tab:table_changes} list all solutions that had some changes made before running the tests. There were four different types of changes made. The most common change was changes to how the solution printed the answer to a problem instance, as the \acp{LLM} often include descriptions in their outputs. Another change regarding the solution outputs was from floating point to integer variables. This made the output of the solutions integers instead of $.0$ vales, aiding parsing in the verification of the output. The change that occurred on multiple solutions was lowered constants to reduce the runtime of the solutions, most common in the P7. These programs did extensive searches on grids larger than needed, causing a long runtime that was cut down by lowering these limits closer to the maximum test values. 

\begin{table}[!ht]
\caption{\label{tab:table_changes} Changes made for LLM generated solutions before running the evaluation tests. Output format indicate changes to the result printing. Float to int indicate the solution used floating point variables when only integer was needed. Lowered constants indicates values being lowered to reduce search sizes. Added imports indicate that some missing imports needed for compilation was added. Lastly main() call indicates a call to the main funtion was added. }
    \centering
    \begin{tabular}{|c|c|c|c|}
    \hline
        \textbf{Problem} & \textbf{LLM} & \textbf{Language} & \textbf{Change} \\ \hline
        P1 & Codepal & C & Output format \\ \hline
        P2 & Llama & C & Float to int \\ \hline
        P2 & Llama & Java  & Float to int \\ \hline
        P2 & Llama & Python & Float to int \\ \hline
        P2 & Mistral & Python & Float to int \\ \hline
        P3 & Copilot & C & Output format \\ \hline
        P3 & Copilot & Python & Output format \\ \hline
        P3 & CodePal & C & Output format \\ \hline
        P3 & CodePal & Java & Output format \\ \hline
        P3 & CodePal & Python & Output format \\ \hline
        P3 & Mistral & Python & main() call \\ \hline
        P7 & Copilot & C & Lowerd constants \\ \hline
        P7 & Codepal & C & Lowerd constants \\ \hline
        P7 & Llama & C & Lowerd constants \\ \hline
        P7 & Llama & Java & Lowerd constants \\ \hline
        P7 & Llama & Python & Lowerd constants \\ \hline
        P9 & Copilot & C & Lowerd constants \\ \hline
        P9 & Mistral & Java & Added import \\ \hline
        P9 & Llama & Python & Output format \\ \hline
        P10 & CodePal & C & Output format \\ \hline
    \end{tabular}
\end{table}

\end{document}